\documentclass{bmvc2k}

\title{\hspace{20pt} Frame-wise Motion and Appearance for \\\hspace{24.5pt}Real-time Multiple Object Tracking}

\addauthor{\hspace{24.5pt}Jimuyang Zhang}{zhangjim@andrew.cmu.edu}{1}
\addauthor{\hspace{24.5pt}Sanping Zhou}{sanpingzhou@stu.xjtu.edu.cn}{2}
\addauthor{\hspace{24.5pt}Jinjun Wang}{jinjun@mail.xjtu.edu.cn}{2}
\addauthor{\hspace{24.5pt}Dong Huang}{donghuang@cmu.edu}{1}

\addinstitution{
 The Robotics Institute,\\
 Carnegie Mellon University,\\
 Pittsburgh, USA
}
\addinstitution{
 Artificial Intelligence and Robotics,\\
 Xi'an Jiaotong University,\\
 Xi'an, China
}

\runninghead{Zhang et al.}{FMA for Real-time MOT}

\graphicspath{{Imgs/}}
\usepackage{times}
\usepackage{epsfig}
\usepackage{graphicx}
\usepackage{amsmath}
\usepackage{amssymb}
\usepackage{color}
\usepackage{multirow}
\usepackage{amssymb}
\usepackage{algorithm}
\usepackage{algorithmic}
\usepackage{array}
\begin{document}

\maketitle

\begin{abstract}
The main challenge of Multiple Object Tracking~(MOT) is the efficiency in associating indefinite number of objects between video frames. Standard motion estimators used in tracking, e.g., Long Short Term Memory~(LSTM), only deal with single object, while Re-IDentification~(Re-ID) based approaches exhaustively compare object appearances. Both approaches are computationally costly when they are scaled to a large number of objects, making it very difficult for real-time MOT. To address these problems, we propose a highly efficient Deep Neural Network~(DNN) that simultaneously models association among indefinite number of objects. The inference computation of the DNN does not increase with the number of objects. Our approach, Frame-wise Motion and Appearance~(FMA), computes the Frame-wise Motion Fields~(FMF) between two frames, which leads to very fast and reliable matching among a large number of object bounding boxes. As auxiliary information is used to fix uncertain matches, Frame-wise Appearance Features~(FAF) are learned in parallel with FMFs. Extensive experiments on the MOT17 benchmark show that our method achieved real-time MOT with competitive results as the state-of-the-art approaches.\footnote{The first two authors contribute equally.}

\end{abstract}

%-------------------------------------------------------------------------
\section{Introduction}
\vspace{-0.2cm}
\label{sec:intro}
The goal of Multiple Object Tracking~(MOT) is to jointly estimate the trajectories of all interested object targets in videos~\cite{Zhang_Wang_Wang:2015,Wang_Fowlkes:2015,Yoon_Boragule_Song:2018,Milan_Rezatofighi_Dick:2017}. MOT has been a critical perception technique required in many real-time applications such as autonomous driving~\cite{Chen_Seff_Koenhauser:2015}, field robotic~\cite{Ross_English_Ball:2015} and video surveillance~\cite{Zhou_Wang_Wang:2017}. 

With the significant achievements made in object detection~\cite{Ren_He_Girshick:2015,Liu_Anguelov_Erhan:2016,Redmon_Divvala_Girshick:2016}, the tracking by detection framework has been popularized in the past few years. Given the object detection results in each frame, this line of tracking methods aim to associate the same objects across different frames. Typically, the object association problem is solved in two steps: (1) Solving a bipartite matching problem between the objects detected in two frames; (2) Optimizing this solution to find one-to-one matches using the Hungarian algorithm~\cite{Kuhn:1955}. In order to achieve high tracking performance, the energy function of bipartite matching is often very computationally costly to solve. As a consequence, the resulting MOT algorithm can hardly be real-time in practical computing platforms, \emph{e.g.}, desktop computing nodes, embedded CPU/GPU modules, and FPGA. 

The situation becomes even worse when the powerful yet computation hungry Deep Neural Networks~(DNN) are used for MOT. DNNs significantly outperform the classical approaches in estimating motion patterns and extracting discriminate appearance features. However, existing DNNs specialized for these tasks cannot be directly used to achieve efficient MOT: (1) These DNNs usually contain heavy back-bone networks for feature extraction from images, such as the ResNet~\cite{He_Zhang_Ren:2016} and GoogleNet~\cite{Szegedy_Liu_Jia:2015}; (2) The off-the-shelf Re-IDentification~(Re-ID) models~\cite{Zhou_Wang_Wang:2017,Sun_Zheng_Yang:2018,Chen_Xu_Li:2018} compute discriminative features independently for each object bounding box, which is very costly when scaling-up to multiple objects; (3) The DNNs that learn object associations~\cite{Sun_Akhtar_Song:2018} need to pre-set the number of objects as a part of the fixed network structure, which cannot handle indefinite number of objects in practice. (4) Time-consuming post-processing algorithms are used to deduce the final association results given appearance and motion patterns~\cite{Tang_Andriluka_Andres:2017} produced by DNNs. As a result, many DNN approaches for MOT do not consider real-time scenarios. 

In this paper, we achieve real-time MOT by using a highly efficient deep learning approach, called Frame-wise Motion and Appearance~(FMA). Our DNN learns (1) Frame-wise Motion Fields~(FMF), a novel association representation that simultaneously accommodates indefinite number of objects. (2) Frame-wise Appearance Feature~(FAF), the holistic discriminative appearance features for Re-ID. Specifically, the pixel-wise responses in FMFs are the coordinate shifts of bounding boxes between two sequential frames. FMFs allow us to use very simple operations to shift bounding boxes and find the corrected matches in both frames. FAFs are learned over all objects of each frame under the Re-ID loss. In inference, they only need to be cropped for a few objects where FMFs are uncertain with one-to-one matches. Besides, we present a simple yet effective inference algorithm to associate object bounding boxes into trajectories. With unoptimized Pytorch (Float32) implementation on a PC with single TitanXP GPU, our FMA has achieved 25FPS on the MOT17 benchmark videos with comparable performance to state-of-the-art algorithms. The main contributions of this work can be highlighted as follows:

\begin{itemize}
    \item Frame-wise Motion Fields~(FMF) to represent the association among indefinite number of objects between frames.
    \item Frame-wise Apearance Feature~(FAF) to provide Re-ID features to assist FMF-based object association.
    \item A simple yet effective inference algorithm to link the objects according to FMFs, and to fix a few uncertain associations using FAFs.
    \item Experiments on the challenging MOT17 benchmark show that our method achieves real-time MOT with competitive performance as the state-of-the-art approaches.
\end{itemize}

%The rest of the paper is organized as follows: Section~\ref{sec:relat} reviews some related works. Our method is introduced in Section~\ref{sec_deep}. Experimental results and ablation studies are presented in Section~\ref{sec:exper}, and the conclusion comes in Section~\ref{sec_concl}. 

\section{Related Work}
\vspace{-0.2cm}
\label{sec:relat}
Because our method jointly learns the motion patterns and appearance features for MOT, we briefly review these two lines of MOT methods, \emph{i.e.}, motion models and appearance models, in the following paragraphs.

\textbf{Motion Model}. The motion models describe how each target moves from frame to frame, which is the key of MOT to locate the search regions of a target in the next frame. Existing motion models can be roughly divided into linear and nonlinear cases. The linear methods~\cite{Milan_Schindler_Roth:2016,Oron_Bar_Avidan:2014,Zhu_Yang_Liu:2018} assume objects move with constant velocity across frames, while the nonlinear methods~\cite{Dicle_Camps_Sznaier:2013,Yang_Nevatia:2012} model inconstant velocities. In the earlier works, the Kalman filter~\cite{Kalman:1960} has been widely used to estimate motion in most MOT methods~\cite{Andriyenko_Schindler:2011,Andriyenko_Schindler_Roth:2012,Kamal_Bappy_Farrell:2016}. With the blooming of DNNs, a large number of deep learning based methods have been designed to learn the motion patterns. For example, the Long Short Term Memory~(LSTM) networks~\cite{Milan_Rezatofighi_Dick:2017,Sadeghian_Alahi_Savarese:2017,Kim_Li_Rehg:2018,Yang_Chan:2017} model can describe and predict complex motion patterns of a target over multiple frames. Both the Kalman filter and LSTM based models belong to the nonlinear methods, therefore they are robust in handling occlusions in the long-term tracking. Due to the strong modeling capability of DNN, the deep motion based methods~\cite{Wan_Wang_Zhou:2018,Sadeghian_Alahi_Savarese:2017} have achieved the state-of-the-art results on the public benchmark datasets. However, these motion models only take the object locations as input, therefore can not explore any appearance or contextual information to help predict the objects positions. Our FMFs are motion representations estimated from appearance and contextual information of the video frames.

\begin{figure}[t]
	\centering
	\begin{tabular}{c}
		\includegraphics[height = 3.5cm, width = 12.5cm]{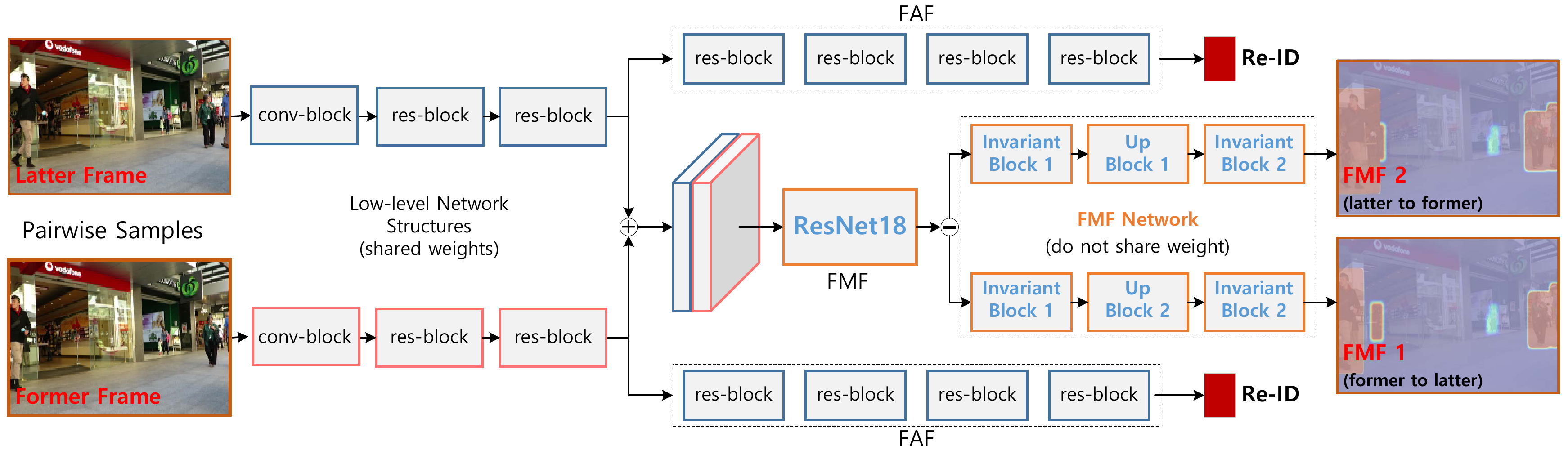}
	\end{tabular}
	\vspace{-0.4cm}
	\caption{DNN architecture of our FMA approach. The DNN takes two frames as input and produces the Frame-wise Motion Fields (FMF) and Frame-wise Re-ID Features~(FAF). FMF simultaneously models the motion of multiple objects, while FAFs provide appearance cues in case FMFs fail to associate certain objects.}
	\label{FMF_network}
	\vspace{-0.2cm}
\end{figure}

\textbf{Appearance Model}. The appearance models produce discriminative features of objects in all frames, such that features of the same object are more similar than those of different objects. In earlier tracking approaches, the color histogram~\cite{Choi_Savarese:2010,Leibe_Schindler_Cornelis:2008,Varior_Shuai_Lu:2016,Le_Heili_Odobez:2016} and pixel-based template representations~\cite{Wu_Thangali_Sclaroff:2012,Pellegrini_Ess_Schindler:2009} are standard hand-crafted features used to describe the appearance of objects. In addition to using Euclidean distances to measure appearance similarity, the covariance matrix representation was also applied to compare the pixel-wise similarities, such as the SIFT-like features~\cite{Fulkerson_Vedaldi_Soatto:2008,Low:2004} and pose features~\cite{Roth_Nevatia_Stiefelhagen:2012,Qin_Shelton:2016}. In recent years, the deep feature learning based methods have been popularized with the blooming of DNNs.  For example, the features of multiple convolution layers are explored to enhance the discriminative capability of learned features~\cite{Chen_Ai_Shang:2017}. In~\cite{Bae_Yoon:2018}, an online feature learning model is designed to associate both the detection results and short tracklets.  Moreover, different network structures, such as siamese network~\cite{Leal_Canton-Ferrer_Schindler:2016}, triplet network~\cite{Ristani_Tomasi:2018} and quadruplet network~\cite{Son_Baek_Cho:2017}, have been extensively used to learn the discriminative features from the detected object bounding boxes. Benefitting from the powerful representation capability of deep features, this line of methods~\cite{Wan_Wang_Kong:2018,Beyer_Breuers_Kurin:2017,Sun_Akhtar_Song:2018,Fernando_Denman_Sridharan:2018} have achieved the state-of-the-art results on the public benchmark datasets. However, most approaches compute discriminative features independently for each object bounding box, which is very costly when scaling-up to multiple objects. Our FAF learning is conducted on the frame basis and multiple objects are handled simultaneously.

\section{Our Method (FMA)}
\vspace{-0.2cm}
\label{sec_deep}

\begin{table}[]
	\centering
	\footnotesize
\begin{tabular}{c|c|c}
\hline
\textbf{Invariant Block 1~($p=1$)} &  \textbf{Up Block}~($s=2$) &  \textbf{Invariant Block 2~($p=1$)}\\ \hline
$\begin{matrix} Conv(3\times3, 512, 512 ) \\ BN + ReLU \\ Conv(3\times3, 512, 512 ) \\ BN + ReLU \end{matrix} $            & $\begin{matrix} Deconv(3\times3, 512, 256) + ReLU \\ Deconv(3\times3, 512, 256) + ReLU \\ Deconv(3\times3, 384, 256) + ReLU \\ Deconv(3\times3, 320, 256) + ReLU \end{matrix} $       & $\begin{matrix} Conv(3\times3, 320, 128) \\ BN + ReLU \\ Conv(3\times3, 128, 64 ) \\ BN + ReLU \\ Conv(3\times3, 64, 2) \end{matrix} $        \\ \hline
\end{tabular}
\caption{The detailed structure of FMF network, in which `Conv' means the normal convolution layer, `BN' denotes the batch normalization layer, `ReLU' indicates the rectified linear unitm, `Deconv' means the normal deconvolution layer, and $p, s$ are the striding and padding parameters, respectively.}
	\label{tab_1}
\end{table}

% \begin{table}[t]
% 	\centering
% 	\footnotesize
% 	\begin{tabular}{  p{3.5cm}<{\centering} |  p{4.5cm}<{\centering} |  p{3.5cm}<{\centering} }
% 		\hline
% 		\textbf{Invariant Block 1~($p=1$)} & \textbf{Up Block}~($s=2$) &  \textbf{Invariant Block 2~($p=1$)}\\
% 		\hline
% 		\multicolumn{1}{c|}{\multirow{5}{*}{\begin{matrix} Conv($3\times3$, 512, 512 ) \\ BN + ReLU \\ Conv($3\times3$, 512, 512 ) \\ BN + ReLU \end{matrix}}} & 
%         \multicolumn{1}{c|}{\multirow{5}{*}{\begin{matrix} Deconv($3\times3$, 512, 256) + ReLU \\ Deconv($3\times3$, 512, 256) + ReLU \\ Deconv($3\times3$, 384, 256) + ReLU \\ Deconv($3\times3$, 320, 256) + ReLU \end{matrix}}}&
%         \multicolumn{1}{c}{\multirow{5}{*}{\begin{matrix} Conv($3\times3$, 320, 128) \\ BN + ReLU \\ Conv($3\times3$, 128, 64 ) \\ BN + ReLU \\ Conv($3\times3$, 64, 2) \end{matrix}}} \\
%         &&\\
%         &&\\
%         &&\\
%         &&\\
% 		\hline
% 	\end{tabular}
% 	\caption{The detailed structure of FMF network, in which `Conv' means the normal convolution layer, `BN' denotes the batch normalization layer, `ReLU' indicates the rectified linear unitm, `Deconv' means the normal deconvolution layer, and $p, s$ are the striding and padding parameters, respectively.}
% 	\label{tab_1}
% \end{table}

Our MOT method consists of a DNN (Figure~\ref{FMF_network}) and an inference algorithm. Candidate objects to be tracked are pre-detected in each frame using the image-based object detectors. For instance, the MOT challenges~\footnote{~\url{https://motchallenge.net/}} provide object candidates detected by DPM~\cite{Felzenszwalb_Girshick_McAllester:2010}, FRCNN~\cite{Ren_He_Girshick:2015} and SDP~\cite{Ess_Leibe_Van:2007}. Our DNN takes two frames as input and produces the Frame-wise Motion Fields (FMFs) and Frame-wise Re-ID Features~(FAFs). The inference algorithm uses FMFs to associate objects between the two frames and uses FAFs to fix associations where FMFs fail. Both the DNN and inference algorithm are very efficient, making it possible for real-time MOT. The DNN consists of conv-blocks and res-blocks of the same structure as~\cite{Wojke_Bewley_Paulus:2017}. Detailed structure of our FMF network is summarized in Table~\ref{tab_1}.

\subsection{Frame-wise Motion Fields}
\begin{figure}[t]
	\centering
	\begin{tabular}{c}
		\includegraphics[height = 4.0cm, width = 12.5cm]{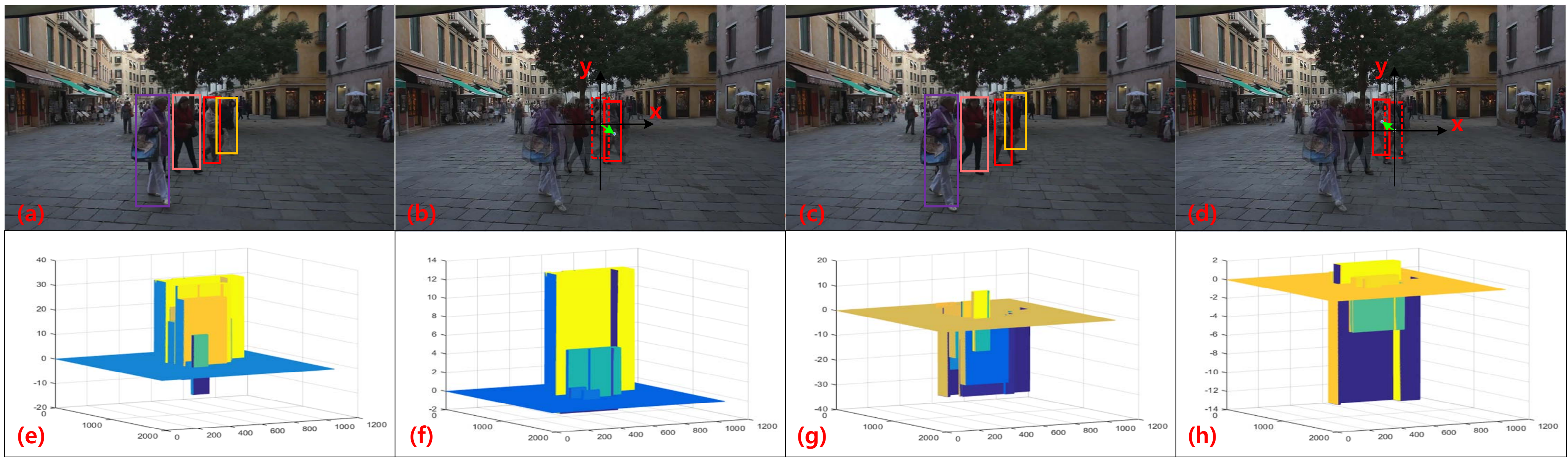}
	\end{tabular}
	\vspace{-0.4cm}
	\caption{Visualization of the FMFs produced on two input frames, in which the colored bounding boxes contain the objects to be tracked. Specifically, (a) and (c) denote two input frames, (b) and (d) illustrate how to decompose the FMFs to horizontal and vertical directions for frame (a) and (c), respectively. The second row visualizes the computed FMFs, in which (e) and (f) represent the horizontal and vertical FMFs for frame (a), while (g) and (h) represent the horizontal and vertical FMFs for frame (c).}
	\label{FMF_label}
	\vspace{-0.2cm}
\end{figure}

Denote the two input frames as the former frame $\mathbf{I}_{1}\in \Re^{w\times h\times 3}$ and the latter frame $\mathbf{I}_{2}\in \Re^{w\times h\times 3}$. Besides, $b_i$ denotes the bounding box of the $i$-th object to be associated between the two frames. The $x$ and $y$ coordinates of $b_i$ in two frames are $\mathcal{X}_1(b_i)$, $\mathcal{X}_2(b_i)$ and $\mathcal{Y}_1(b_i)$, $\mathcal{Y}_2(b_i)$, respectively. The DNN (Figure~\ref{FMF_network}) produces four FMFs that represent the motion between frame $\mathbf{I}_{1}$ and frame $\mathbf{I}_{2}$. Formally, the ground truth of FMFs is computed as:
\begin{eqnarray}
   & \mathbf{F}_x^1(b_i) = \mathcal{X}_2(b_i) - \mathcal{X}_1(b_i), \mathbf{F}_y^1(b_i) = \mathcal{Y}_2(b_i) - \mathcal{Y}_1(b_i),
    \label{eq_1}\\
    & \mathbf{F}_x^2(b_i) = \mathcal{X}_1(b_i) - \mathcal{X}_2(b_i), \mathbf{F}_y^2(b_i) = \mathcal{Y}_1(b_i) - \mathcal{Y}_2(b_i),
    \label{eq_2}
\end{eqnarray}
where $\mathbf{F}_x^1(b_i)\in \Re^{w\times h\times 1}$ and $\mathbf{F}_y^1(b_i) \in \Re^{w\times h\times 1}$ contains the motion vectors of bounding box $b_i$ assigned on the pixels in $\mathbf{I}_{1}$. Similarly, $\mathbf{F}_x^2(b_i)$, $\mathbf{F}_y^2(b_i)$ contains the motion vectors of bounding box $b_i$ assigned on the pixels in $\mathbf{I}_{2}$. In Figure~\ref{FMF_label}, we visualize the FMFs on two input frames. Note that we chose to predict the bi-directional motion vectors of $b_i$ between $\mathbf{I}_{1}$ and $\mathbf{I}_{2}$, respectively. This design enforces robust estimation of motion vectors: since $b_i$ in former and latter frames usually contain different appearance and contextual information, they can help each other in estimating motions. Moreover, it also deals with two cases in inference: \textit{Case (1)} certain $b_i$ exists in $\mathbf{I}_{1}$ but disappears in $\mathbf{I}_{2}$, and \textit{Case (2)} certain $b_i$ is not in $\mathbf{I}_{1}$ but appears in $\mathbf{I}_{2}$. In our later inference algorithm, $\mathbf{F}_x^1(b_i)$ and $\mathbf{F}_y^1(b_i)$ are used to associate the detections from the former to latter frame, while $\mathbf{F}_x^2(b_i)$ and $\mathbf{F}_y^2(b_i)$ are used to associate the detections from the later to former frame.  

The FMFs simultaneously embed multiple object affinity information and can be directly learned by the network in an end-to-end manner. To learn the FMFs, we simply use the Mean Square Error~(MSE) as the loss function:
\begin{equation}
\vspace{-0.1cm}
\mathcal{L}_{MSE} = \sum\limits_{k = 1}^2 \sum\limits_{b_i \in \mathbf{B}}  \|\mathbf{H}_x^k(b_i) - \mathbf{F}_x^k(b_i)\|_F^2 + \|\mathbf{H}_y^k(b_i) - \mathbf{F}_y^k(b_i)\|_F^2, \label{eq_3}
\vspace{-0.1cm}
\end{equation} 
where $\mathbf{H}_y^k$ and $\mathbf{H}_y^k$ are the outputs of network on the former and latter frames, and $\mathbf{B}$ is the set of all the object bounding boxes to be associated.

\subsection{Frame-wise Appearance Features}
Our FMFs can successfully solve association for most objects in benchmark database, but there are still some extreme cases. For instance, when frames are noisy and objects are crowed, FMFs may not be estimated perfectly for every objects between two frames. As a practical extension for extreme cases, our MOT approach finds several candidate bounding boxes according to FMFs, and uses the re-identification~(Re-ID) approach to verify the identity of an object among these candidates. 

The standard Re-ID approaches~\cite{Zhou_Wang_Wang:2017,Sun_Zheng_Yang:2018} mainly focus on generating powerful appearance representation to associate the same object captured from difference cameras and far-away time lapses. They take image patches of each object bounding box and use a specialized DNNs to learn discriminative features from these image patches. Given the powerful FMFs, we found that learning decent Re-ID features only requires a few convolutions. Moreover, to achieve fast MOT, we avoid using an independent DNN for Re-ID feature learning by jointly computing the Re-ID features with FMFs. We also keep the DNN inference computation invariant to the number of objects by learning frame-wise features. As shown in Fig.~\ref{FMF_network}, our DNN learns FAFs in parallel with FMFs. 

Given the bounding boxes of objects in two frames, our Re-ID step crops patches from FAFs and computes the similarity between two cropped features. In training, the cropped features of the same objects are concatenated as the positive samples, and that of different objects are concatenated as negative samples. In practice, the number of negative and positive samples are kept as $4:1$. FMFs are learned by training on the Binary Cross-Entropy~(BCE) loss:
\begin{equation}
\label{eq_4}
\vspace{-0.1cm}
\mathcal{L}_{BCE} = -\frac{1}{N}\sum\limits_{j = 1}^N \mathcal{S}_j \log h(\mathbf{p}_j) + (1 - \mathcal{S}_j) \log(1 - h(\mathbf{p}_j))
\vspace{-0.1cm}
\end{equation} 
where $\mathbf{p}_j$ is the $j$th sample, $\mathcal{S}_j$ denotes the class label of $\mathbf{p}_j$, in which $\mathcal{S}_j = 0$ means $\mathbf{p}_j$ contains features from the same object, otherwise $\mathcal{S}_j = 1$.

\begin{algorithm}[!t]
    \footnotesize
	\caption{Inference algorithm based on FMFs and FAFs for real-time MOT}
	\label{alg}
	\begin{algorithmic}[1]
		\STATE {\bfseries Input:} Initial detections: $\mathbf{D} = \{\mathbf{D}_i\}_{i=1}^N$, final FMF model: $\mathbf{H}$ and Re-ID model $\mathbf{P}$.
		\STATE {\bfseries Parameter:} $\tau_1$  (IOU threshold) and $\tau_2$ (Re-ID similarity threshold).
		\STATE {\bfseries Output:} Final tracks: $\mathbf{T} = \{\mathbf{T}^{k}\}_{k=1}^M$, where $\mathbf{T}^k  = \{\mathbf{t}_i^{k}\}_{i=1}^N$.
		\STATE Initialize the tracks $\mathbf{T}$ using $\mathbf{D}_1$;
		\FOR{$i=1$; $i<N$; $i+\hspace{-0.05cm}+$}
		\STATE Initialize the track $\mathbf{T}_A$, $\mathbf{T}_B$, $\mathbf{T}_C$ and $\mathbf{T}_D$ using $\mathbf{T}$, $a = 0$ and $b=0$;
		
		\FOR{$k = 1$; $k \leq \|\mathbf{T}_A^i\|$; $k+\hspace{-0.05cm}+$}  
		\STATE Compute the IOU score: $\mathbf{d}_{i+1}^k, \mathbf{s}_{i+1}^k = \mathrm{IOU}(\mathbf{H}_{1,2}(\mathbf{T}_A^{i,k}), \mathbf{D}_{i+1})$;
		\STATE Find the candidates: $ \mathbf{s}_c = \mathrm{SORT}(\mathbf{s}_{i+1}^k > \tau_1)$
		\IF{$\|\mathbf{s}_c\| = 1$}
		\STATE Update $\mathbf{t}_i^k$ by adding $\mathbf{d}_{i+1}^k(1)$, $\mathbf{D}_{i+1}$ by deleting $\mathbf{d}_{i+1}^k(1)$, $\mathbf{T}_B^i$, $\mathbf{T}_C^i$ and $\mathbf{T}_D^i$ by deleting $\mathbf{t}_i^k$;
		\STATE ++a;
        \ELSE
        \STATE Compute the appearance score: $d_{i+1}^k, c_{i+1}^k = \max\{\mathrm{SIM}(\mathbf{P}(\mathbf{T}_A^{i,k}), \mathbf{P}(\mathbf{D}_{i+1}))\}$;
        \IF{$c_{i+1}^k > \tau_2$}
        \STATE Update $\mathbf{t}_i^k$ by adding $d_{i+1}^k$, $\mathbf{D}_{i+1}$ by deleting $d_{i+1}^k$, $\mathbf{T}_B^i$, $\mathbf{T}_C^i$ and $\mathbf{T}_D^i$ by deleting $\mathbf{t}_i^k$;
        \STATE ++a;
        \ELSE
        \STATE Update $\mathbf{t}_i^k$ by adding $\mathbf{d}_{i+1}^k(1)$, $\mathbf{D}_{i+1}$ by deleting $\mathbf{d}_{i+1}^k(1)$, $\mathbf{T}_B^i$, $\mathbf{T}_C^i$ and $\mathbf{T}_D^i$ by deleting $\mathbf{t}_i^k$;
        \STATE ++a;
        \ENDIF
		\ENDIF
        \ENDFOR
        
        \FOR{$k = 1$; $k \leq \|\mathbf{T}_B^i\|$; $k+\hspace{-0.05cm}+$}
        \STATE Compute the IOU score: $\mathbf{d}_{i+1}^{k+a}, \mathbf{s}_{i+1}^{k+a} = \mathrm{IOU}(\mathbf{H}_{3,4}(\mathbf{D}_{i+1}), \mathbf{T}_B^{i,k+a})$;
        \STATE Find the candidates: $ \mathbf{s}_c = \mathrm{SORT}(\mathbf{s}_{i+1}^{k+a} > \tau_1)$
        \IF{$\|\mathbf{s}_c\| = 1$}
		\STATE Update $\mathbf{t}_i^{k+a}$ by adding $\mathbf{d}_{i+1}^{k+a}(1)$, $\mathbf{D}_{i+1}$ by deleting $\mathbf{d}_{i+1}^{k+a}(1)$, $\mathbf{T}_C^i$ and $\mathbf{T}_D^i$ by deleting $\mathbf{t}_i^{k+a}$;
		\STATE ++b;
        \ELSE
        \STATE Compute the appearance score: $d_{i+1}^{k+a}, c_{i+1}^{k+a} = \max\{\mathrm{SIM}(\mathbf{P}(\mathbf{T}_B^{i,k+a}), \mathbf{P}(\mathbf{D}_{i+1}))\}$;
        \IF{$c_{i+1}^{k+a} > \tau_2$}
        \STATE Update $\mathbf{t}_i^{k+a}$ by adding $d_{i+1}^{k+a}$, $\mathbf{D}_{i+1}$ by deleting $d_{i+1}^k$, $\mathbf{T}_C^i$ and $\mathbf{T}_D^i$ by deleting $\mathbf{t}_i^{k+a}$;
        \STATE ++b;
        \ELSE
        \STATE Update $\mathbf{t}_i^{k+a}$ by adding $\mathbf{d}_{i+1}^{k+a}(1)$, $\mathbf{D}_{i+1}$ by deleting $\mathbf{d}_{i+1}^{k+a}(1)$, $\mathbf{T}_C^i$ and $\mathbf{T}_D^i$ by deleting $\mathbf{t}_i^{k+a}$;
		\STATE ++b;
        \ENDIF
        \ENDIF
        \ENDFOR

        \FOR{$k = 1$; $k \leq \|\mathbf{T}_C^i\|$; $k+\hspace{-0.05cm}+$}
        \STATE Compute the appearance score: $d_{i+1}^{k+a+b}, c_{i+1}^{k+a+b} = \max\{\mathrm{SIM}(\mathbf{P}(\mathbf{T}_C^{i,k+a+b}), \mathbf{P}(\mathbf{D}_{i+1}))\}$;
        \IF{$c_{i+1}^{k+a+b} > \tau_2$}
        \STATE Update $\mathbf{t}_i^{k+a+b}$ by adding $d_{i+1}^{k+a+b}$, $\mathbf{D}_{i+1}$ by deleting $d_{i+1}^k$, and $\mathbf{T}_D^i$ by deleting $\mathbf{t}_i^{k+a+b}$;
        \ENDIF
        \ENDFOR
        \STATE Initialize the rest of $\mathbf{D}_{i+1}$ as new targets and terminate the targets in $\mathbf{T}_D^i$.
		\ENDFOR
	\end{algorithmic}
\end{algorithm}

\subsection{Inference Algorithm}
Given FMFs and FAFs, our inference algorithm efficiently associates the detection results between frames, and produces the track for each detected object. 

Formally, $\mathbf{D} = \{\mathbf{D}_1, \dots, \mathbf{D}_N\}$ denotes the detection results, \emph{e.g.}, object bounding boxes, in $N$ videos frames. $\mathbf{T}=\{\mathbf{T}^{1},\dots, \mathbf{T}^{M}\}$ denotes the tracks of $M$ objects. $\mathrm{IOU}(\cdot)$ is the operator to compute the bounding box overlap between the last detection of an active track and candidate detections in the next frame, and $\mathrm{SIM}(\cdot)$ measures the appearance similarities between the last detection of an active track and candidate detections in the next frame. Our inference algorithm conducts MOT in three steps, \textbf{\textcolor{blue}{Step 1}}: Associate the tracks from the former to latter frame; \textbf{\textcolor{blue}{Step 2}}: Associate the tracks from the later to former frame; \textbf{\textcolor{blue}{Step 3}}: Associate the remaining tracks and detections with FAFs. For simplicity, we use $\mathbf{T}_A$, $\mathbf{T}_B$, $\mathbf{T}_C$, $\mathbf{T}_D$ to store the unmatched tracks after each step. The detailed inference process is summarized in Algorithm~\ref{alg}. The algorithm does not need to solve any optimization problem or learn any parameters. We only introduce two thresholds $\tau_1$  (IOU threshold) and $\tau_2$ (Re-ID similarity threshold) to help decide the association process. 

\section{Experiments}
\vspace{-0.2cm}
\label{sec:exper}

\subsection{Dataset and Setting}
\hspace{0.45cm}
\textbf{Dataset}. We conducted experiments on the MOT17 dataset, which is the latest benchmark for multiple human tracking. This dataset contains various challenging video sequences recorded by static or moving cameras, and under the complex scenes of illumination changes, varying viewpoints and weather conditions. In total, there are $7$ fully annotated training videos and $7$ testing videos, in which public detection results are obtained by three image-based object detectors: DPM~\cite{Felzenszwalb_Girshick_McAllester:2010}, FRCNN~\cite{Ren_He_Girshick:2015} and SDP~\cite{Ess_Leibe_Van:2007}.

\textbf{Setting}. In the training process, the input frame-pairs were composed under two intervals: two frames every $1$ frame, and every $4$ frames. The training batch size is $4$, the learning rate is initialized to be $0.001$ and decreased by a factor of $0.1$ at $70$ epochs. In the testing process, the input frame-pairs are taken under as two frames every $1$ frame. The inference hardware is a PC with an Intel Xeon(R) $3.80$GHz CPU, $64$GB RAM, and a NVIDIA GTX TitanXP GPU. The DNN was implemented in Pytorch under Float32 and the inference algorithm in Python without parallel speedup. We set $\tau_1 = 0.45$ and $\tau_2 = 0.5$ for all the testing videos.

\subsection{Ablation Study}
\vspace{-0.2cm}
In this section, we analyze our method from two aspects: MOT result of each component and effectiveness of FMFs.

\textbf{MOT result of each component}. Our DNN jointly learns FMFs and FAFs for MOT. We show in Table~\ref{tab_2} that when conducting inference using only FMFs or FAFs, the performances are close to each other but very different in speed. FMFs can effectively associate the detections between two frames, resulting in very fast MOT at $27.5$ Hz. Without FMFs, inference with only FAFs need to conduct Re-ID on every pair of candidate bounding boxes, leading to a much slower speed at $5.3$ Hz. In FMA~(FMF+FAF), FMFs can find high-quality matches for most objects, and FAFs were only used to handle the remaining hard cases. As a result, FMA shows good trade-off between speed and accuracy.  

\begin{table*}[!t]
	\footnotesize
	\centering
	\begin{tabular}{| p{2cm}<{\centering} | p{1.0cm}<{\centering} | p{0.8cm}<{\centering} | p{0.8cm}<{\centering} | p{0.8cm}<{\centering} | p{0.9cm}<{\centering} | p{0.6cm}<{\centering} | p{0.8cm}<{\centering} | p{0.8cm}<{\centering} |}
		\hline
		\textbf{Methods} & MOTA $\uparrow$ & MT $\uparrow$  & ML $\downarrow$ & FP $\downarrow$ &
		FN $\downarrow$    &
		ID$\downarrow$ &
		Frag  $\downarrow$ & Hz  $\uparrow$\\
		\hline
		\textbf{FMF}  & 56.3 & 21.5\% & 33.2\% & \textbf{4,961} & 74,754 & 2,539 & \textbf{3,141}  & 27.5\\
		\textbf{FAF}  & 57.3 & 23.8\% & 30.6\% & 6,632 & \textbf{71,707} & 2,012 & 5,083   & 5.3 \\
		\textbf{FMA~(FMF+FAF)}  & \textbf{57.5} & \textbf{24.1}\% & \textbf{30.2}\% & 6,655 & 71,720 & \textbf{1,553} & 4,899 & 25.2\\
		\hline
	\end{tabular}
	\caption{Results achieved by each component of our method on the MOT17 benchmark.}
	\label{tab_2}
	\vspace{-0.2cm}
\end{table*}

\begin{table*}[!t]
	\footnotesize
	\centering
	\begin{tabular}{| p{1.0cm}<{\centering} | p{1.0cm}<{\centering} | p{1.0cm}<{\centering} | p{0.7cm}<{\centering} | p{0.7cm}<{\centering} | p{0.8cm}<{\centering} | p{0.9cm}<{\centering} | p{0.6cm}<{\centering} | p{0.8cm}<{\centering} | p{0.6cm}<{\centering} |}
		\hline
		\textbf{Methods} & MOTA $\uparrow$ & MOTP $\uparrow$ & MT $\uparrow$  & ML $\downarrow$ & FP $\downarrow$ &
		FN $\downarrow$    &
		ID$\downarrow$ &
		Frag  $\downarrow$ &
		Hz $\uparrow$ \\
		\hline
		PHD  & 48.0 & 77.2 & 17.1\% & 35.6\% & 23,199 & 265,954 & 3,998 & 8,886 & 6.7 \\
		EAMTT & 42.6 & 76.0 & 12.7\% & 42.7\% & 30,711 & 288,474 & 4,488 & 5,720 & 1.4 \\
        GMPHD & 39.6 & 74.5 & 8.8\% & 43.3\% & 50,903 & 284,228 & 5,811 & 7,414 & 3.3 \\
        GM\_PHD & 36.4 & 74.5 & 4.1\% & 57.3\% & 23,723 & 330,767 & 4,607 & 11,317 & 38.4\\
        MTDF    & 49.6 & 75.5 & 18.9\% & 33.1\% & 37,124 & 241,768 & 5,567 & 9,260 & 1.2\\
        \hline
        MOTDT & 50.9 & 76.6 & 17.5\% & 35.7\% & 24,069 & 250,768 & 2,474 & 5,317 & 18.3\\
        \textbf{FMA} & 47.4 & 76.6 & 17.1\% & 35.8\% & 21,498 & 271,237 & 4,019 & 13,107 & 25.2\\
		\hline
	\end{tabular}
	\caption{Comparisons with state-of-the-art online methods on the MOT17 benchmark. MOT metrics were computed over all three object detectors provided by MOT17.}
	\label{tab_3}
	\vspace{-0.2cm}
\end{table*}

\begin{figure}[!t]
	\centering
	\begin{tabular}{c}
		\includegraphics[height = 3.5cm, width = 12.5cm]{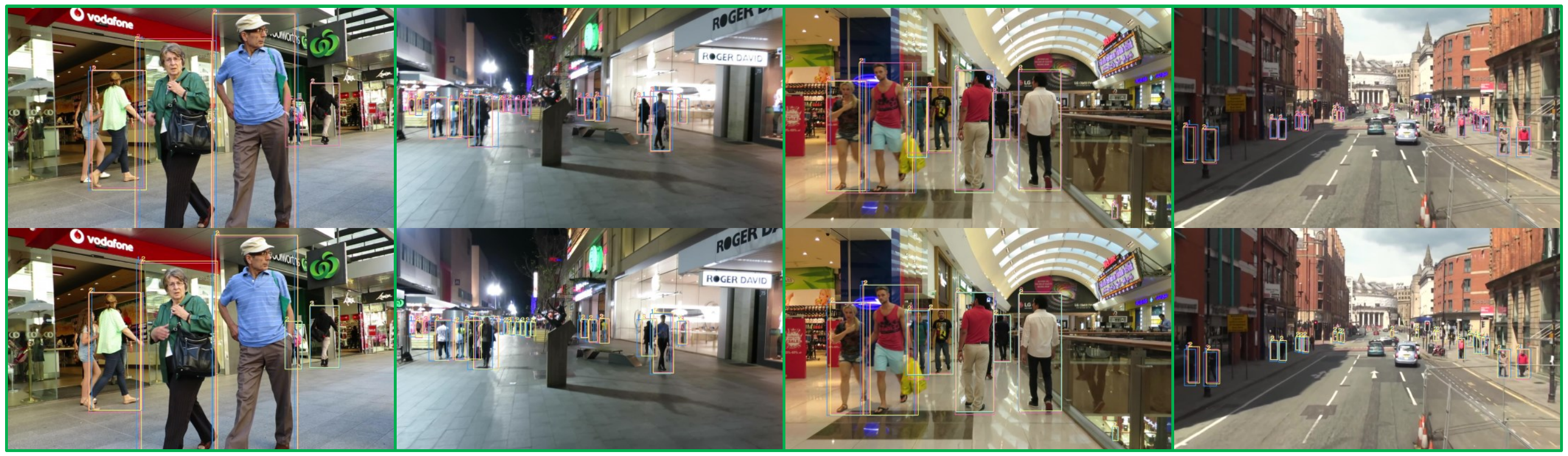}
	\end{tabular}
	\vspace{-0.4cm}
	\caption{The predicted bounding boxes (yellow) by FMFs between two adjacent frames. The blue and magenta boxes are the bounding boxes provided by the SDP object detectors.}
	\label{FMF_visual}
	\vspace{-0.4cm}
\end{figure}

\begin{table*}[t]
	\footnotesize
	\centering
	\begin{tabular}{| p{1.5cm}<{\centering} | p{1.0cm}<{\centering} | p{1.0cm}<{\centering} | p{0.8cm}<{\centering} | p{0.8cm}<{\centering} | p{0.9cm}<{\centering} | p{0.8cm}<{\centering} | p{0.8cm}<{\centering} | p{0.8cm}<{\centering} |}
		\hline
		\textbf{Detectors} & \textbf{Methods} & MOTA $\uparrow$ & MT $\uparrow$  & ML $\downarrow$ & FP $\downarrow$ &
		FN $\downarrow$    &
		ID$\downarrow$ &
		Frag  $\downarrow$ \\
		\hline
		\multicolumn{1}{|c|}{\multirow{3}{*}{DPM~\cite{Felzenszwalb_Girshick_McAllester:2010}}} & MOTDT         & 46.5\% & 14.6\% & 40.5\% & 8,841 & 90,990 & 802 & 1,860 \\
        &MTDF          & 44.5\% & 13.6\% & 39.0\% & 11,822 & 90,539 & 1,999 & 3,371 \\
        &\textbf{FMA} & 39.6\% & 10.8\% & 43.7\% & 5,409 & 106,965 & 1,276 & 5,065 \\
		\hline
		\multicolumn{1}{|c|}{\multirow{3}{*}{FRCNN~\cite{Ren_He_Girshick:2015}}} & MOTDT         & 47.7\% &  15.4\% & 35.9\% & 8,911 & 88,773 & 731 & 1,540 \\
        &MTDF          & 47.2\% &  18.7\% & 31.8\% & 15,119 & 82,331 & 1,804 & 2,937 \\
        &\textbf{FMA} & 45.1\% &  16.3\% & 33.6\% & 9,434 & 92,552 & 1,190 & 3,143 \\
        \hline
        \multicolumn{1}{|c|}{\multirow{3}{*}{SDP~\cite{Ess_Leibe_Van:2007}}} &MOTDT         & 58.4\% &  22.5\% & 30.7\% & 6,317 & 71,005 & 941 & 1,917 \\
        &MTDF          & 57.0\% &  24.2\% & 28.4\% & 10,183 & 68,898 & 1,764 & 2,952 \\
        &\textbf{FMA} & 57.5\% &  24.1\% & 30.2\% & 6,655 & 71,720 & 1,553 & 4,899 \\
        \hline
	\end{tabular}
	\caption{Comparisons of MOT methods with individual detectors on the MOT17 benchmark.}
	\label{tab_4}
	\vspace{-0.2cm}
\end{table*}

\textbf{Effectiveness of FMFs}. FMFs contain pixel-wise estimation of displacement between the detected bounding boxes in adjacent frames: a former frame and a latter frame. From FMFs, our algorithm computes the displacement for each bounding box, and produces a prediction of each bounding box in both the former and latter frame. The one-to-one matching between the predicted bounding box and the detected bounding box is simply found by computing IOU. Figure~\ref{FMF_visual} shows that our approach can effectively deal with the complex motion patterns in the tracking process. The first row and the second row are the former and the latter frames, respectively. The the predicted bounding boxes are plotted in yellow, and the bounding boxes in the former and latter frame are plotted in blue and magenta, respectively. The high IOUs between the predicted and detected bounding boxes indicate that the learned FMFs can successfully associate the detections between two adjacent frames.

\subsection{Comparison}
\vspace{-0.2cm}
In this section, we compare our method with other state-of-the-art online methods on the MOT17 challenge benchmark. The compared methods include the MOTDT~\cite{Chen_Ai_Zhuang:2018}, PHD~\cite{Fu_Feng_Angelini:2018}, EAMTT~\cite{Sanchez_Poiesi_Cavallaro:2016}, GMPHD~\cite{Kutschbach_Bochinski_Eiselein:2017}, GM\_PHD~\cite{Eiselein_Arp_Michael:2012}, MTDF~\cite{Fu_Angelini_Chambers:2019} models, in which the MOTDT is a deep learning based method and the others are based on traditional approaches. 

The results are shown in Table~\ref{tab_3}, from which we can conclude that: (1) The deep learning based methods usually outperform the traditional methods in accuracy. (2) Our method has achieved a faster speed while performs a bit worse in overall accuracy than MOTDT. \textbf{Note that}, FMA achieved 25FPS with unoptimized Pytorch (Float32) implementation on a PC with single-TitanXP GPU. The above performance was achieved only by training from scratch on the MOT17 data.

The reason for the lower overall accuracy of our approach comes from two aspects: (1) Unlike MOTDT that used external person Re-ID datasets (\emph{i.e.}, CUHK01~\cite{Li_Zhao_Wang:2012}, CUHK03~\cite{Li_Zhao_Xiao:2014}, Market1501~\cite{Zheng_Shen_Tian:2015}), our method was only trained on MOT17 data and did not use any external data to enhance FAFs; (2) Our method favors strong detection results, because accurate FMFs and FAFs need to be learned from scratch in training. In Table~\ref{tab_4}, we show MOT performance using three different detector results provided by MOT17. Here we compare our method with two best-performing models, \emph{i.e.}, MOTDT and MTDF. We can see that our method achieves very competitive results when the FRCNN and SDP detectors were used.

\begin{figure}[!t]
	\centering
	\begin{tabular}{c}
		\includegraphics[height = 5.5cm, width = 12.5cm]{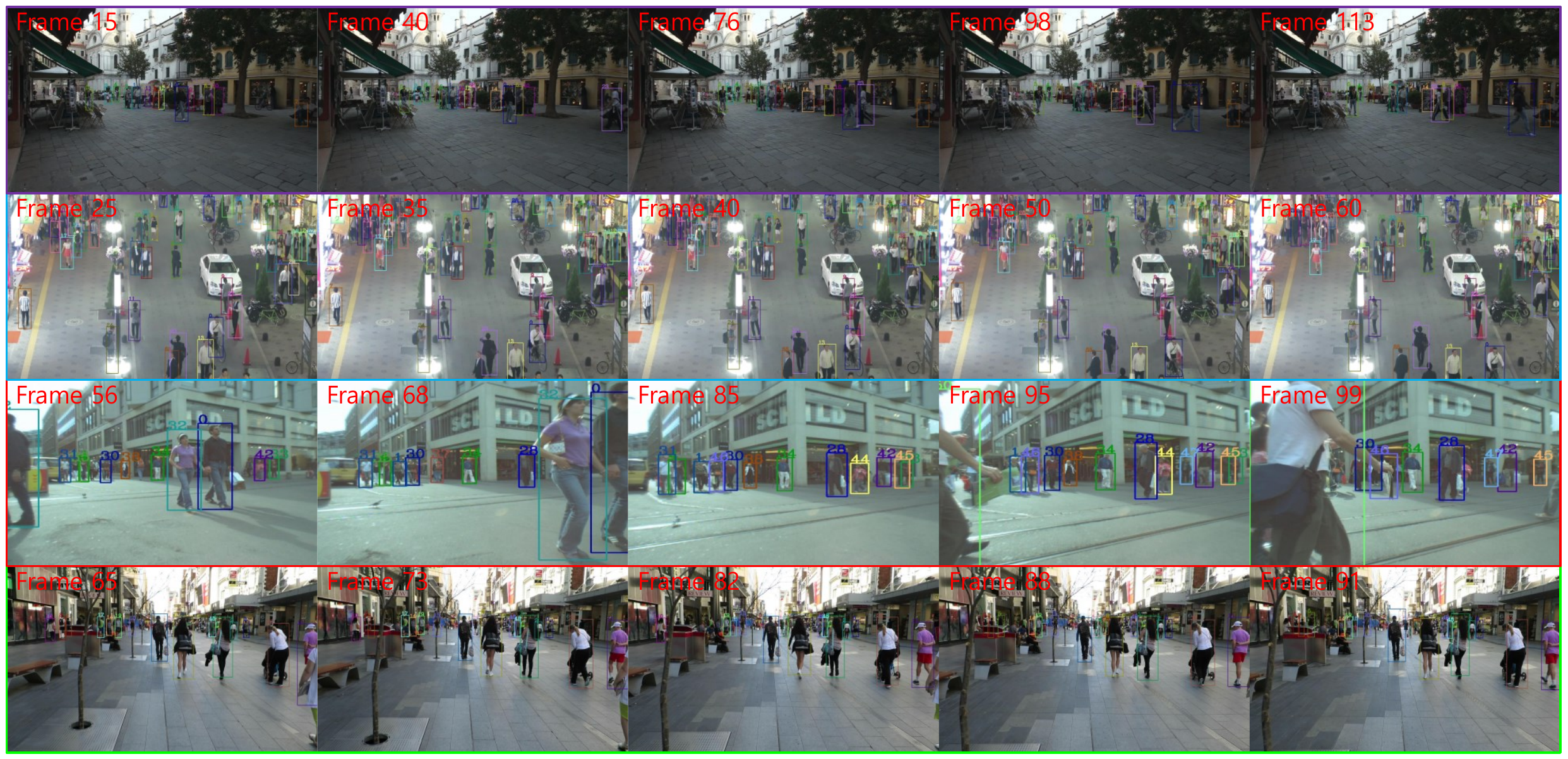}
	\end{tabular}
	\vspace{-0.4cm}
	\caption{Challenging tracking examples by FMA with the SDP detector. Different targets are marked by different colors and numbers.}
	\label{FMF_example}
	\vspace{-0.4cm}
\end{figure}

Moreover, we show some tracking examples by our method on four challenging video sequences in Figure~\ref{FMF_example}. Our method is robust to illumination changes, varying viewpoints and mutual occlusion. Nevertheless, in challenging videos, it is not easy to get accurate frame-based object detection results. In the future, we plan to further relief the reliance on strong detectors by training FMFs on short video clips instead of two frames. This will make FMF robust to outliers in the frame-wise detection results.

\section{Conclusion}
\vspace{-0.2cm}
\label{sec_concl}

Practical and real-time MOT has to scale well with indefinite number of objects. This paper addresses this problem with frame-wise representations of object motion and appearance. In particular, FMFs simultaneously handle forward and backward motions of all bounding boxes in two input frames, which is the key to achieve real-time MOT inference. FAFs helps FMFs in handling some hard cases without significantly compromising the speed. The FMFs and FAFs are efficiently used in our inference algorithm, and achieved faster and more competitive results on the MOT17 benchmark. Our frame-wise representations are very efficient and general, making it possible to achieve real-time inference on more computationally expensive tracking tasks, such as instance segmentation tracking and scene mapping.

{
\small
\bibliography{FMF}

\begin{thebibliography}{60}
\providecommand{\natexlab}[1]{#1}
\providecommand{\url}[1]{\texttt{#1}}
\expandafter\ifx\csname urlstyle\endcsname\relax
  \providecommand{\doi}[1]{doi: #1}\else
  \providecommand{\doi}{doi: \begingroup \urlstyle{rm}\Url}\fi

\bibitem[Wan(2018)]{Wan_Wang_Kong:2018}
Multi-object tracking using online metric learning with long short-term memory.
\newblock In \emph{ICIP}, pages 788--792. IEEE, 2018.

\bibitem[Andriyenko and Schindler(2011)]{Andriyenko_Schindler:2011}
Anton Andriyenko and Konrad Schindler.
\newblock Multi-target tracking by continuous energy minimization.
\newblock In \emph{CVPR}, pages 1265--1272. IEEE, 2011.

\bibitem[Andriyenko et~al.(2012)Andriyenko, Schindler, and
  Roth]{Andriyenko_Schindler_Roth:2012}
Anton Andriyenko, Konrad Schindler, and Stefan Roth.
\newblock Discrete-continuous optimization for multi-target tracking.
\newblock In \emph{CVPR}, pages 1926--1933. IEEE, 2012.

\bibitem[Bae and Yoon(2018)]{Bae_Yoon:2018}
Seung-Hwan Bae and Kuk-Jin Yoon.
\newblock Confidence-based data association and discriminative deep appearance
  learning for robust online multi-object tracking.
\newblock \emph{TPAMI}, 40\penalty0 (3):\penalty0 595--610, 2018.

\bibitem[Beyer et~al.(2017)Beyer, Breuers, Kurin, and
  Leibe]{Beyer_Breuers_Kurin:2017}
Lucas Beyer, Stefan Breuers, Vitaly Kurin, and Bastian Leibe.
\newblock Towards a principled integration of multi-camera re-identification
  and tracking through optimal bayes filters.
\newblock In \emph{CVPR Workshops}, pages 29--38, 2017.

\bibitem[Chen et~al.(2015)Chen, Seff, Kornhauser, and
  Xiao]{Chen_Seff_Koenhauser:2015}
Chenyi Chen, Ari Seff, Alain Kornhauser, and Jianxiong Xiao.
\newblock Deepdriving: Learning affordance for direct perception in autonomous
  driving.
\newblock In \emph{ICCV}, pages 2722--2730, 2015.

\bibitem[Chen et~al.(2018)Chen, Xu, Li, Sebe, and Wang]{Chen_Xu_Li:2018}
Dapeng Chen, Dan Xu, Hongsheng Li, Nicu Sebe, and Xiaogang Wang.
\newblock Group consistent similarity learning via deep crf for person
  re-identification.
\newblock In \emph{CVPR}, pages 8649--8658, 2018.

\bibitem[Chen et~al.(2017)Chen, Ai, Shang, Zhuang, and Bai]{Chen_Ai_Shang:2017}
Long Chen, Haizhou Ai, Chong Shang, Zijie Zhuang, and Bo~Bai.
\newblock Online multi-object tracking with convolutional neural networks.
\newblock In \emph{ICIP}, pages 645--649. IEEE, 2017.

\bibitem[Choi and Savarese(2010)]{Choi_Savarese:2010}
Wongun Choi and Silvio Savarese.
\newblock Multiple target tracking in world coordinate with single, minimally
  calibrated camera.
\newblock In \emph{ECCV}, pages 553--567. Springer, 2010.

\bibitem[Dicle et~al.(2013)Dicle, Camps, and Sznaier]{Dicle_Camps_Sznaier:2013}
Caglayan Dicle, Octavia~I Camps, and Mario Sznaier.
\newblock The way they move: Tracking multiple targets with similar appearance.
\newblock In \emph{ICCV}, pages 2304--2311, 2013.

\bibitem[Eiselein et~al.(2012)Eiselein, Arp, P{\"a}tzold, and
  Sikora]{Eiselein_Arp_Michael:2012}
Volker Eiselein, Daniel Arp, Michael P{\"a}tzold, and Thomas Sikora.
\newblock Real-time multi-human tracking using a probability hypothesis density
  filter and multiple detectors.
\newblock In \emph{2012 IEEE Ninth International Conference on Advanced Video
  and Signal-Based Surveillance}, pages 325--330. IEEE, 2012.

\bibitem[Ess et~al.(2007)Ess, Leibe, and Van~Gool]{Ess_Leibe_Van:2007}
Andreas Ess, Bastian Leibe, and Luc Van~Gool.
\newblock Depth and appearance for mobile scene analysis.
\newblock In \emph{ICCV}, pages 1--8. IEEE, 2007.

\bibitem[Felzenszwalb et~al.(2010)Felzenszwalb, Girshick, McAllester, and
  Ramanan]{Felzenszwalb_Girshick_McAllester:2010}
Pedro~F Felzenszwalb, Ross~B Girshick, David McAllester, and Deva Ramanan.
\newblock Object detection with discriminatively trained part-based models.
\newblock \emph{TPAMI}, 32\penalty0 (9):\penalty0 1627--1645, 2010.

\bibitem[Fernando et~al.(2018)Fernando, Denman, Sridharan, and
  Fookes]{Fernando_Denman_Sridharan:2018}
Tharindu Fernando, Simon Denman, Sridha Sridharan, and Clinton Fookes.
\newblock Tracking by prediction: A deep generative model for mutli-person
  localisation and tracking.
\newblock In \emph{WACV}, pages 1122--1132. IEEE, 2018.

\bibitem[Fu et~al.(2018)Fu, Feng, Angelini, Chambers, and
  Naqvi]{Fu_Feng_Angelini:2018}
Zeyu Fu, Pengming Feng, Federico Angelini, Jonathon Chambers, and Syed~Mohsen
  Naqvi.
\newblock Particle phd filter based multiple human tracking using online
  group-structured dictionary learning.
\newblock \emph{IEEE Access}, 6:\penalty0 14764--14778, 2018.

\bibitem[Fu et~al.(2019)Fu, Angelini, Chambers, and
  Naqvi]{Fu_Angelini_Chambers:2019}
Zeyu Fu, Federico Angelini, Jonathon Chambers, and Syed~Mohsen Naqvi.
\newblock Multi-level cooperative fusion of gm-phd filters for online multiple
  human tracking.
\newblock \emph{TMM}, 2019.

\bibitem[Fulkerson et~al.(2008)Fulkerson, Vedaldi, and
  Soatto]{Fulkerson_Vedaldi_Soatto:2008}
Brian Fulkerson, Andrea Vedaldi, and Stefano Soatto.
\newblock Localizing objects with smart dictionaries.
\newblock In \emph{ECCV}, pages 179--192. Springer, 2008.

\bibitem[He et~al.(2016)He, Zhang, Ren, and Sun]{He_Zhang_Ren:2016}
Kaiming He, Xiangyu Zhang, Shaoqing Ren, and Jian Sun.
\newblock Deep residual learning for image recognition.
\newblock In \emph{CVPR}, pages 770--778, 2016.

\bibitem[Kalman(1960)]{Kalman:1960}
Rudolph~Emil Kalman.
\newblock A new approach to linear filtering and prediction problems.
\newblock \emph{Journal of basic Engineering}, 82\penalty0 (1):\penalty0
  35--45, 1960.

\bibitem[Kamal et~al.(2016)Kamal, Bappy, Farrell, and
  Roy-Chowdhury]{Kamal_Bappy_Farrell:2016}
Ahmed~T Kamal, Jawadul~H Bappy, Jay~A Farrell, and Amit~K Roy-Chowdhury.
\newblock Distributed multi-target tracking and data association in vision
  networks.
\newblock \emph{TPAMI}, 38\penalty0 (7):\penalty0 1397--1410, 2016.

\bibitem[Kim et~al.(2018)Kim, Li, and Rehg]{Kim_Li_Rehg:2018}
Chanho Kim, Fuxin Li, and James~M Rehg.
\newblock Multi-object tracking with neural gating using bilinear lstm.
\newblock In \emph{ECCV}, pages 200--215, 2018.

\bibitem[Kuhn(1955)]{Kuhn:1955}
Harold~W Kuhn.
\newblock The hungarian method for the assignment problem.
\newblock \emph{Naval research logistics quarterly}, 2\penalty0 (1-2):\penalty0
  83--97, 1955.

\bibitem[Kutschbach et~al.(2017)Kutschbach, Bochinski, Eiselein, and
  Sikora]{Kutschbach_Bochinski_Eiselein:2017}
Tino Kutschbach, Erik Bochinski, Volker Eiselein, and Thomas Sikora.
\newblock Sequential sensor fusion combining probability hypothesis density and
  kernelized correlation filters for multi-object tracking in video data.
\newblock In \emph{AVSS}, pages 1--5. IEEE, 2017.

\bibitem[Le et~al.(2016)Le, Heili, and Odobez]{Le_Heili_Odobez:2016}
Nam Le, Alexander Heili, and Jean-Marc Odobez.
\newblock Long-term time-sensitive costs for crf-based tracking by detection.
\newblock In \emph{ECCV}, pages 43--51. Springer, 2016.

\bibitem[Leal-Taix{\'e} et~al.(2016)Leal-Taix{\'e}, Canton-Ferrer, and
  Schindler]{Leal_Canton-Ferrer_Schindler:2016}
Laura Leal-Taix{\'e}, Cristian Canton-Ferrer, and Konrad Schindler.
\newblock Learning by tracking: Siamese cnn for robust target association.
\newblock In \emph{CVPR Workshops}, pages 33--40, 2016.

\bibitem[Leibe et~al.(2008)Leibe, Schindler, Cornelis, and
  Van~Gool]{Leibe_Schindler_Cornelis:2008}
Bastian Leibe, Konrad Schindler, Nico Cornelis, and Luc Van~Gool.
\newblock Coupled object detection and tracking from static cameras and moving
  vehicles.
\newblock \emph{TMAMI}, 30\penalty0 (10):\penalty0 1683--1698, 2008.

\bibitem[Li et~al.(2012)Li, Zhao, and Wang]{Li_Zhao_Wang:2012}
Wei Li, Rui Zhao, and Xiaogang Wang.
\newblock Human reidentification with transferred metric learning.
\newblock In \emph{ACCV}, 2012.

\bibitem[Li et~al.(2014)Li, Zhao, Xiao, and Wang]{Li_Zhao_Xiao:2014}
Wei Li, Rui Zhao, Tong Xiao, and Xiaogang Wang.
\newblock Deepreid: Deep filter pairing neural network for person
  re-identification.
\newblock In \emph{CVPR}, 2014.

\bibitem[Liu et~al.(2016)Liu, Anguelov, Erhan, Szegedy, Reed, Fu, and
  Berg]{Liu_Anguelov_Erhan:2016}
Wei Liu, Dragomir Anguelov, Dumitru Erhan, Christian Szegedy, Scott Reed,
  Cheng-Yang Fu, and Alexander~C Berg.
\newblock Ssd: Single shot multibox detector.
\newblock In \emph{ECCV}, pages 21--37. Springer, 2016.

\bibitem[Long et~al.(2018)Long, Haizhou, Zijie, and Chong]{Chen_Ai_Zhuang:2018}
Chen Long, Ai~Haizhou, Zhuang Zijie, and Shang Chong.
\newblock Real-time multiple people tracking with deeply learned candidate
  selection and person re-identification.
\newblock In \emph{ICME}, 2018.

\bibitem[Lowe(2004)]{Low:2004}
David~G Lowe.
\newblock Distinctive image features from scale-invariant keypoints.
\newblock \emph{IJCV}, 60\penalty0 (2):\penalty0 91--110, 2004.

\bibitem[Milan et~al.(2016)Milan, Schindler, and
  Roth]{Milan_Schindler_Roth:2016}
Anton Milan, Konrad Schindler, and Stefan Roth.
\newblock Multi-target tracking by discrete-continuous energy minimization.
\newblock \emph{TPAMI}, 38\penalty0 (10):\penalty0 2054--2068, 2016.

\bibitem[Milan et~al.(2017)Milan, Rezatofighi, Dick, Reid, and
  Schindler]{Milan_Rezatofighi_Dick:2017}
Anton Milan, S~Hamid Rezatofighi, Anthony Dick, Ian Reid, and Konrad Schindler.
\newblock Online multi-target tracking using recurrent neural networks.
\newblock In \emph{AAAI}, 2017.

\bibitem[Oron et~al.(2014)Oron, Bar-Hille, and Avidan]{Oron_Bar_Avidan:2014}
Shaul Oron, Aharon Bar-Hille, and Shai Avidan.
\newblock Extended lucas-kanade tracking.
\newblock In \emph{ECCV}, pages 142--156. Springer, 2014.

\bibitem[Pellegrini et~al.(2009)Pellegrini, Ess, Schindler, and
  Van~Gool]{Pellegrini_Ess_Schindler:2009}
Stefano Pellegrini, Andreas Ess, Konrad Schindler, and Luc Van~Gool.
\newblock You'll never walk alone: Modeling social behavior for multi-target
  tracking.
\newblock In \emph{ICCV}, pages 261--268. IEEE, 2009.

\bibitem[Qin and Shelton(2016)]{Qin_Shelton:2016}
Zhen Qin and Christian~R Shelton.
\newblock Social grouping for multi-target tracking and head pose estimation in
  video.
\newblock \emph{TMAMI}, 38\penalty0 (10):\penalty0 2082--2095, 2016.

\bibitem[Redmon et~al.(2016)Redmon, Divvala, Girshick, and
  Farhadi]{Redmon_Divvala_Girshick:2016}
Joseph Redmon, Santosh Divvala, Ross Girshick, and Ali Farhadi.
\newblock You only look once: Unified, real-time object detection.
\newblock In \emph{CVPR}, pages 779--788, 2016.

\bibitem[Ren et~al.(2015)Ren, He, Girshick, and Sun]{Ren_He_Girshick:2015}
Shaoqing Ren, Kaiming He, Ross Girshick, and Jian Sun.
\newblock Faster r-cnn: Towards real-time object detection with region proposal
  networks.
\newblock In \emph{NIPS}, pages 91--99, 2015.

\bibitem[Ristani and Tomasi(2018)]{Ristani_Tomasi:2018}
Ergys Ristani and Carlo Tomasi.
\newblock Features for multi-target multi-camera tracking and
  re-identification.
\newblock In \emph{Proceedings of the IEEE Conference on Computer Vision and
  Pattern Recognition}, pages 6036--6046, 2018.

\bibitem[Ross et~al.(2015)Ross, English, Ball, Upcroft, and
  Corke]{Ross_English_Ball:2015}
Patrick Ross, Andrew English, David Ball, Ben Upcroft, and Peter Corke.
\newblock Online novelty-based visual obstacle detection for field robotics.
\newblock In \emph{ICRA}, pages 3935--3940. IEEE, 2015.

\bibitem[Roth et~al.(2012)Roth, B{\"a}uml, Nevatia, and
  Stiefelhagen]{Roth_Nevatia_Stiefelhagen:2012}
Markus Roth, Martin B{\"a}uml, Ram Nevatia, and Rainer Stiefelhagen.
\newblock Robust multi-pose face tracking by multi-stage tracklet association.
\newblock In \emph{ICPR}, pages 1012--1016. IEEE, 2012.

\bibitem[Sadeghian et~al.(2017)Sadeghian, Alahi, and
  Savarese]{Sadeghian_Alahi_Savarese:2017}
Amir Sadeghian, Alexandre Alahi, and Silvio Savarese.
\newblock Tracking the untrackable: Learning to track multiple cues with
  long-term dependencies.
\newblock In \emph{ICCV}, pages 300--311, 2017.

\bibitem[Sanchez-Matilla et~al.(2016)Sanchez-Matilla, Poiesi, and
  Cavallaro]{Sanchez_Poiesi_Cavallaro:2016}
Ricardo Sanchez-Matilla, Fabio Poiesi, and Andrea Cavallaro.
\newblock Online multi-target tracking with strong and weak detections.
\newblock In \emph{ECCV}, pages 84--99. Springer, 2016.

\bibitem[Son et~al.(2017)Son, Baek, Cho, and Han]{Son_Baek_Cho:2017}
Jeany Son, Mooyeol Baek, Minsu Cho, and Bohyung Han.
\newblock Multi-object tracking with quadruplet convolutional neural networks.
\newblock In \emph{ICCV}, pages 5620--5629, 2017.

\bibitem[Sun et~al.(2018{\natexlab{a}})Sun, Akhtar, Song, Mian, and
  Shah]{Sun_Akhtar_Song:2018}
ShiJie Sun, Naveed Akhtar, HuanSheng Song, Ajmal Mian, and Mubarak Shah.
\newblock Deep affinity network for multiple object tracking.
\newblock \emph{arXiv preprint arXiv:1810.11780}, 2018{\natexlab{a}}.

\bibitem[Sun et~al.(2018{\natexlab{b}})Sun, Zheng, Yang, Tian, and
  Wang]{Sun_Zheng_Yang:2018}
Yifan Sun, Liang Zheng, Yi~Yang, Qi~Tian, and Shengjin Wang.
\newblock Beyond part models: Person retrieval with refined part pooling (and a
  strong convolutional baseline).
\newblock In \emph{ECCV}, pages 480--496, 2018{\natexlab{b}}.

\bibitem[Szegedy et~al.(2015)Szegedy, Liu, Jia, Sermanet, Reed, Anguelov,
  Erhan, Vanhoucke, and Rabinovich]{Szegedy_Liu_Jia:2015}
Christian Szegedy, Wei Liu, Yangqing Jia, Pierre Sermanet, Scott Reed, Dragomir
  Anguelov, Dumitru Erhan, Vincent Vanhoucke, and Andrew Rabinovich.
\newblock Going deeper with convolutions.
\newblock In \emph{CVPR}, pages 1--9, 2015.

\bibitem[Tang et~al.(2017)Tang, Andriluka, Andres, and
  Schiele]{Tang_Andriluka_Andres:2017}
Siyu Tang, Mykhaylo Andriluka, Bjoern Andres, and Bernt Schiele.
\newblock Multiple people tracking by lifted multicut and person
  re-identification.
\newblock In \emph{CVPR}, pages 3539--3548, 2017.

\bibitem[Varior et~al.(2016)Varior, Shuai, Lu, Xu, and
  Wang]{Varior_Shuai_Lu:2016}
Rahul~Rama Varior, Bing Shuai, Jiwen Lu, Dong Xu, and Gang Wang.
\newblock A siamese long short-term memory architecture for human
  re-identification.
\newblock In \emph{ECCV}, pages 135--153. Springer, 2016.

\bibitem[Wan et~al.(2018)Wan, Wang, and Zhou]{Wan_Wang_Zhou:2018}
Xingyu Wan, Jinjun Wang, and Sanping Zhou.
\newblock An online and flexible multi-object tracking framework using long
  short-term memory.
\newblock In \emph{CVPR Workshops}, pages 1230--1238, 2018.

\bibitem[Wang and Fowlkes(2015)]{Wang_Fowlkes:2015}
Shaofei Wang and Charless~C Fowlkes.
\newblock Learning optimal parameters for multi-target tracking.
\newblock In \emph{BMVC}, volume~1, page~6, 2015.

\bibitem[Wojke et~al.(2017)Wojke, Bewley, and Paulus]{Wojke_Bewley_Paulus:2017}
Nicolai Wojke, Alex Bewley, and Dietrich Paulus.
\newblock Simple online and realtime tracking with a deep association metric.
\newblock In \emph{ICIP}, pages 3645--3649. IEEE, 2017.

\bibitem[Wu et~al.(2012)Wu, Thangali, Sclaroff, and
  Betke]{Wu_Thangali_Sclaroff:2012}
Zheng Wu, Ashwin Thangali, Stan Sclaroff, and Margrit Betke.
\newblock Coupling detection and data association for multiple object tracking.
\newblock In \emph{CVPR}, pages 1948--1955. IEEE, 2012.

\bibitem[Yang and Nevatia(2012)]{Yang_Nevatia:2012}
Bo~Yang and Ram Nevatia.
\newblock An online learned crf model for multi-target tracking.
\newblock In \emph{CVPR}, pages 2034--2041. IEEE, 2012.

\bibitem[Yang and Chan(2017)]{Yang_Chan:2017}
Tianyu Yang and Antoni~B Chan.
\newblock Recurrent filter learning for visual tracking.
\newblock In \emph{ICCV}, pages 2010--2019, 2017.

\bibitem[Yoon et~al.(2018)Yoon, Boragule, Song, Yoon, and
  Jeon]{Yoon_Boragule_Song:2018}
Young-chul Yoon, Abhijeet Boragule, Young-min Song, Kwangjin Yoon, and Moongu
  Jeon.
\newblock Online multi-object tracking with historical appearance matching and
  scene adaptive detection filtering.
\newblock In \emph{AVSS}, pages 1--6. IEEE, 2018.

\bibitem[Zhang et~al.(2015)Zhang, Wang, Wang, Gong, and
  Liu]{Zhang_Wang_Wang:2015}
Shun Zhang, Jinjun Wang, Zelun Wang, Yihong Gong, and Yuehu Liu.
\newblock Multi-target tracking by learning local-to-global trajectory models.
\newblock \emph{Pattern Recognition}, 48\penalty0 (2):\penalty0 580--590, 2015.

\bibitem[Zheng et~al.(2015)Zheng, Shen, Tian, Wang, Wang, and
  Tian]{Zheng_Shen_Tian:2015}
Liang Zheng, Liyue Shen, Lu~Tian, Shengjin Wang, Jingdong Wang, and Qi~Tian.
\newblock Scalable person re-identification: A benchmark.
\newblock In \emph{ICCV}, 2015.

\bibitem[Zhou et~al.(2017)Zhou, Wang, Wang, Gong, and
  Zheng]{Zhou_Wang_Wang:2017}
Sanping Zhou, Jinjun Wang, Jiayun Wang, Yihong Gong, and Nanning Zheng.
\newblock Point to set similarity based deep feature learning for person
  re-identification.
\newblock In \emph{CVPR}, pages 3741--3750, 2017.

\bibitem[Zhu et~al.(2018)Zhu, Yang, Liu, Kim, Zhang, and
  Yang]{Zhu_Yang_Liu:2018}
Ji~Zhu, Hua Yang, Nian Liu, Minyoung Kim, Wenjun Zhang, and Ming-Hsuan Yang.
\newblock Online multi-object tracking with dual matching attention networks.
\newblock In \emph{ECCV}, pages 366--382, 2018.

\end{thebibliography}
}
\end{document}